%% file: main.tex
\documentclass[10pt,twocolumn,letterpaper]{article}

\usepackage[table,xcdraw,dvipsnames]{xcolor}
\usepackage{cvpr}              % To produce the CAMERA-READY version

\usepackage{float}


\definecolor{cvprblue}{rgb}{0.21,0.49,0.74}
\definecolor{softred}{RGB}{200, 90, 80}
\definecolor{LightOrange}{rgb}{1.0, 0.9, 0.8}
\definecolor{DeeperOrange}{rgb}{1.0, 0.85, 0.7}
\definecolor{LightGreen}{rgb}{0.8, 1.0, 0.8}

\usepackage[pagebackref,breaklinks,colorlinks,allcolors=cvprblue]{hyperref}

\def\paperID{0000} % *** Enter the Paper ID here
\def\confName{CVPR}
\def\confYear{2025}

\title{PAINT: Paying Attention to INformed Tokens to Mitigate Hallucination in Large Vision-Language Model}

\author{
Kazi Hasan Ibn Arif, Sajib Acharjee Dip, Khizar Hussain, Lang Zhang, Chris Thomas\\
Virginia Tech, Blacksburg, Virginia, USA\\
{\tt\small \{hasanarif, sajibacharjeedip, khizar, langzhang, chris\}@vt.edu}
}

\begin{document}    

\maketitle

\begin{abstract}

Large Vision Language Models (LVLMs) have demonstrated remarkable capabilities in understanding and describing visual content, achieving state-of-the-art performance across various vision-language tasks. However, these models often generate descriptions containing objects or details that are absent in the input image, a phenomenon commonly known as hallucination. Our work investigates the key reasons behind this issue by analyzing the pattern of self-attention in transformer layers. We find that hallucinations often arise from the progressive weakening of attention weight to visual tokens in the deeper layers of the LLM. Some previous works naively boost the attention of all visual tokens to mitigate this issue, resulting in suboptimal hallucination reduction. To address this, we identify two critical sets of visual tokens that facilitate the transfer of visual information from the vision encoder to the LLM. Local tokens encode grounded information about objects present in an image, while summary tokens capture the overall aggregated representation of the image. Importantly, these two sets of tokens require different levels of weight enhancement. To this end, we propose \textbf{PAINT} (\textbf{P}aying \textbf{A}ttention to \textbf{IN}formed \textbf{T}okens), a plug-and-play framework that intervenes in the self-attention mechanism of the LLM, selectively boosting the attention weights of local and summary tokens with experimentally learned margins. Evaluation on the MSCOCO image captioning dataset demonstrate that our approach reduces hallucination rates by up to 62.3\% compared to baseline models while maintaining accuracy. Code is available at \href{https://github.com/hasanar1f/PAINT}{https://github.com/hasanar1f/PAINT}

\end{abstract}

\input{sec/1_intro}

\input{sec/2_related_work}
\input{sec/3_method}
\input{sec/4_exp}
\input{sec/6_conclusion}

{
    \small
    \bibliographystyle{ieeenat_fullname}
    \bibliography{main}
}

\end{document}

%% file: sec/1_intro.tex
\section{Introduction}
\label{sec:intro}

Large Vision Language Models (LVLMs) attract considerable interest due to advancements in pre-trained models that bridge visual and textual embedding spaces \cite{radford2021learning, jia2021scaling, shu2024visualtextmeetslowlevel, Hong_2024_CVPR, wang2023visionllmlargelanguagemodel, dai2023instructblipgeneralpurposevisionlanguagemodels}. This focus spans both new architectural designs \cite{gao2023llama, zhu2023minigpt, huang2023visual, liu2024visual, laurençon2024buildingbetterunderstandingvisionlanguage, 10445007, LIAO2024100116, huang2024surveyevaluationmultimodallarge} and the creation of comprehensive benchmarking datasets \cite{lu2023mathvista, xu2023lvlm, li2023vlmevalgeneralevaluationvideo}. Similar to text only LLMs, LVLMs experience hallucination generating descriptions for nonexistent objects in images, which affects accuracy in critical areas like medical imaging \cite{yin2023survey, wang2023evaluation, lovenia2023negative, liu2024surveyhallucinationlargevisionlanguage, li2024referencefreehallucinationdetectionlarge}.

% \begin{figure}[H]
%     \centering
%     \includegraphics[width=0.99\linewidth]{llava-hallu.pdf}
%     \caption{(a) An example of in-context hallucination in LLaVA-1.5~\cite{llava}. The responses that are not grounded in the image are highlighted in \textcolor{softred}{color}. (b) Visualization of the embeddings of visual tokens and text tokens in the semantic space, along with the full token vocabulary of Vicuna-7B. The figure clearly shows that the visual tokens, projected by the MLP into the text embedding space, are significantly distant from the text token embeddings, indicating a modality gap.}
%     \label{fig:hallu-exmp}
% \end{figure}

\begin{figure}[H]
    \centering
    \includegraphics[width=0.99\linewidth]{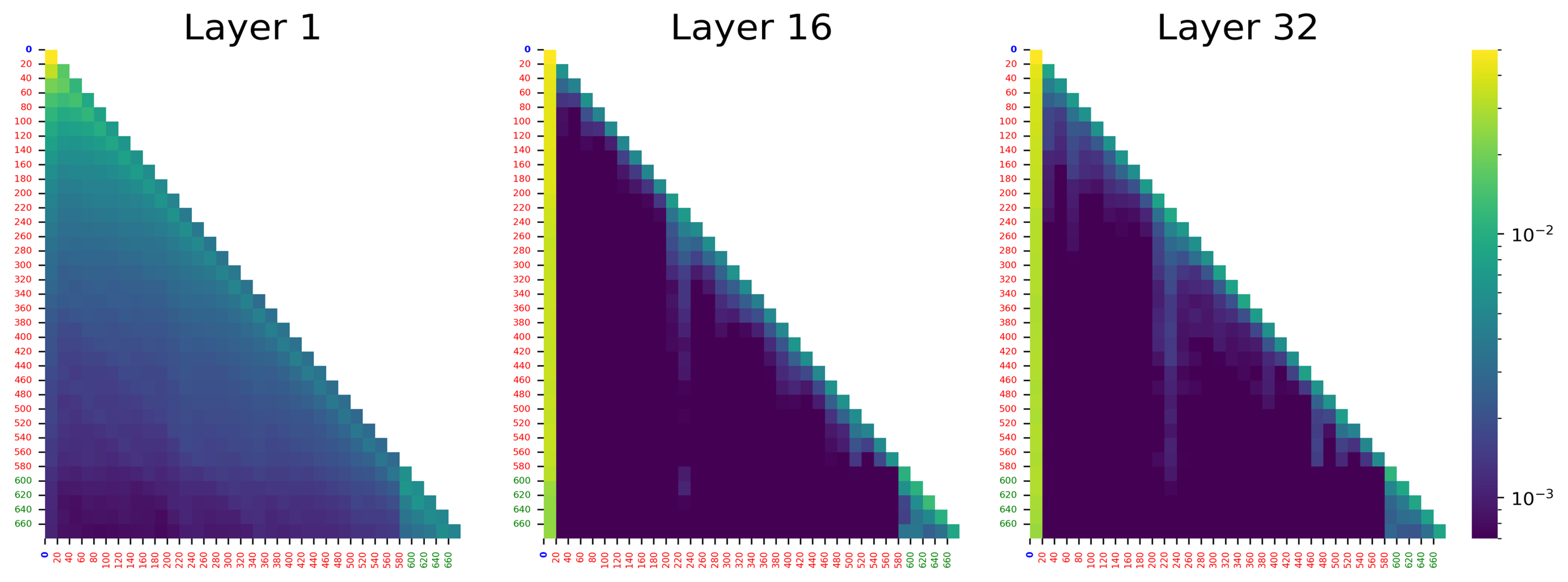}
    \caption{Imbalanced self-attention weights observed in LLaVA-1.5. Image tokens (labeled in \textcolor{red}{red}) receive lower attention weights than text tokens, especially in shallow layers (e.g., 16, 32).}
    \label{fig:visualize-attn}
\end{figure}

A typical LVLM consists of a pretrained Vision Transformer (ViT)~\cite{vit,siglip} and a pretrained Large Language Model (LLM)~\cite{vicuna,opt,qwen} with a lightweight MLP projector. First, the input image is transformed into visual tokens through the ViT and project layer. Then, the LLM process the visual and text tokens together to generate the response. LVLMs often generate plausible but incorrect information about the image, commonly known as hallucination. To investigate this phenomena, we  visualize the attention values (before applying softmax) for both visual and text tokens during the generation process, as shown in Fig~\ref{fig:visualize-attn}. The results show that in shallow layers (e.g., Layer 1), attention is distributed relatively uniformly across both visual and text tokens, suggesting initial comprehensive processing of all input modalities. In intermediate layers (e.g., Layer 16), the model begins to show preferential attention to specific token clusters, particularly focusing on summary tokens that capture high-level semantic information. Finally, in deeper layers (e.g., Layer 32), attention becomes heavily concentrated on a small subset of tokens, primarily text-based summary tokens, while attention to visual tokens diminishes significantly. In summary, we observe a progressive weakening of attention weights in LVLMs. This pattern contributes to the hallucination observed in current MLLMs~\cite{opera}.

To mitigate hallucination, a recent work, PAI \cite{liu2024payingattentionimagetrainingfree} suggests to boost up the attention weights of visual tokens in LLM layers. However, this uniform attention enhancement leads to suboptimal hallucination reduction. Our investigation reveals that there exist two critical sets of visual tokens that facilitate the transfer of visual information from the vision encoder to the LLM. Local tokens encode grounded information about objects present in an image, while summary tokens capture the overall aggregated representation of the image. Importantly, these two sets of tokens require different levels of attention enhancement. To this end, we propose \textbf{PAINT} (\textbf{P}aying \textbf{A}ttention to \textbf{IN}formed \textbf{T}okens), a plug-and-play self-attention modification in the LLM, that selectively boosts the attention weights of top $N$ local and summary tokens with some margins $\alpha$ and $\beta$. The value of the margins as well as $N$ are identified through our experiments. 

We evaluated our method on the MSCOCO~\cite{mscoco} dataset, demonstrating that our approach reduces hallucination rates by up to 62.3\% compared to the previous method, PAI~\cite{liu2024payingattentionimagetrainingfree}, while maintaining strong task performance. In summary, we make three major contributions in this study. \textbf{(1)}~Through experiments on vision encoders, we identify local tokens that store grounded information and summary tokens that store aggregated information from images, which are then transferred to the LLM. \textbf{(2)}~We propose a plug-and-play technique that requires no additional training and intervenes in the self-attention layers of LVLMs, boosting the attention of local and summary tokens with learned margins to mitigate hallucination. \textbf{(3)}~We evaluate our method on standard hallucination benchmarks, demonstrating that it outperforms state-of-the-art approaches.

%% file: sec/2_related_work.tex
\section{Previous Works}
\label{sec:related_works}

Many recent studies have explored various techniques to mitigate hallucinations in Large Vision Language Models (LVLMs)~\cite{ma2024vista,li2023evaluatingobjecthallucinationlarge, sarkar2024mitigatingobjecthallucinationdata, kim2024codecontrastingselfgenerateddescription, yu2024hallucidoctormitigatinghallucinatorytoxicity, su2024mitigatingentitylevelhallucinationlarge, zhu2024ibdalleviatinghallucinationslarge, liu2024payingattentionimagetrainingfree, huang2024operaalleviatinghallucinationmultimodal, an2024aglamitigatingobjecthallucinations, favero2024multimodalhallucinationcontrolvisual}. Among them, the PAI (Paying Attention to Image)~\cite{liu2024payingattentionimagetrainingfree} stands out for its simple yet effective hallucination mitigation. At its core, PAI~\cite{liu2024payingattentionimagetrainingfree} uniformly enhance the self-attention weights of visual tokens by a constant factor $\alpha$. Additionally, PAI~\cite{liu2024payingattentionimagetrainingfree} employs logit refinement to balance visual and textual contributions. While PAI~\cite{liu2024payingattentionimagetrainingfree} improves upon previous baselines, it has several limitations. The primary issue is that it applies a uniform boost to all visual tokens, disregarding their individual significance. This indiscriminate enhancement can amplify noise and degrade model performance. Additionally, the static enhancement factor $\alpha$ does not adapt to different contexts, and the framework treats visual tokens as independent entities, ignoring spatial and semantic relationships.

%% file: sec/3_method.tex
\section{Methodology}
\label{sec:methods}
%------------------------------------------------------------------------

\begin{figure}
    \centering
    \includegraphics[width=0.99\linewidth]{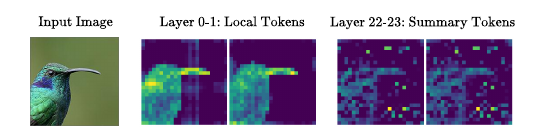}
    \caption{CLS-to-patch attention in ViT~\cite{vit}. Initial layers (e.g., 0-1) highlight local patches containing fine-grained visual details (\textit{local tokens}), while deeper layers (e.g., 22-23) emphasize summary tokens that aggregate global information (\textit{summary tokens}).}
    \label{fig:vit-cls-attn}
\end{figure}

\begin{figure}
    \centering
    \includegraphics[width=0.99\linewidth]{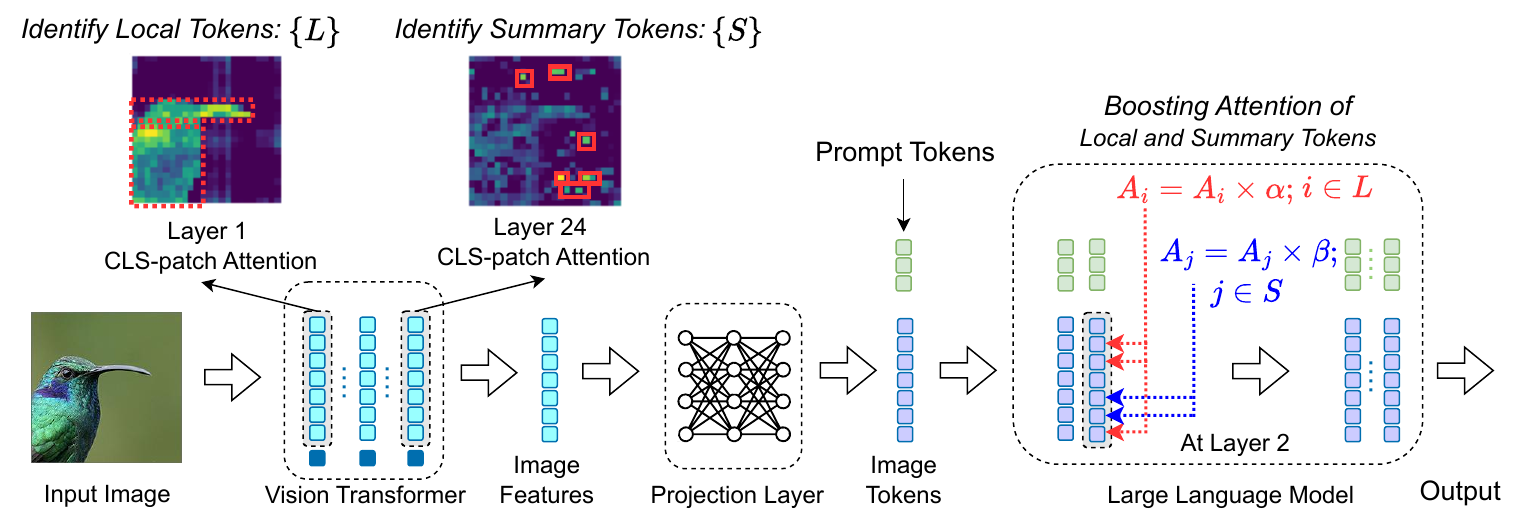}
    \caption{PAINT hallucination mitigation method}
    \label{fig:paint}
\end{figure}

We propose Paying Attention to INformed Tokens (PAINT), a hallucination mitigation method. Unlike previous methods that uniformly treat all visual tokens the same, we identify two kinds of visual tokens, known as local and summary tokens (detailed in section~\ref{subsec:attention_patterns}), and boost their attention with $\alpha$ and $\beta$, respectively, as discussed in section~\ref{subsec:our_approach}. PAINT is illustrated in Fig~\ref{fig:paint}.

\subsection{Local and Summary Tokens}
\label{subsec:attention_patterns}
Inspired by previous studies~\cite{vit-register,hired}, we observe a distinct phenomenon in how ViT encodes the input image. The initial layers of ViT primarily capture local information from image patches, referred to as \textit{local tokens}. In contrast, deeper layers aggregate this local information to extract global representations, storing them in background tokens that were initially less informative. These tokens, termed \textit{summary tokens}, play a crucial role in global information retention~\cite{huang2024operaalleviatinghallucinationmultimodal}.

This behavior can be attributed to the hierarchical integration of visual information, where summary tokens hold global semantics while local tokens retain fine-grained details. This phenomenon is evident in the CLS-to-patch attention maps of ViTs at different depths, as shown in Fig~\ref{fig:vit-cls-attn}. In the early layers, attention is concentrated on local patches, directly capturing grounded details, whereas in deeper layers, the attention shifts towards summary tokens that encode higher-level semantics. For instance, in the ‘bird’ image in Fig~\ref{fig:vit-cls-attn}, local tokens focus on patches corresponding to the bird’s eyes, beak, and feathers, encoding grounded visual details. Conversely, deeper layers prioritize summary tokens, which encapsulate abstract semantics such as the bird’s species and action.

\subsection{Key Insights}

The presence of both local and summary tokens is critical in the decoding stage of large vision-language models (LVLMs). Current LVLM architectures position vision tokens at the beginning of the sequence, expecting them to contribute to a precise understanding of the image. However, as text generation progresses, vision-related information may become attenuated due to interactions between local and summary tokens. We hypothesize that local and summary tokens require different levels of attention enhancement, rather than the uniform boosting applied in previous methods. The appropriate degree of enhancement remains an open research question, which we explore through further experiments. To address this, we introduce two hyperparameters, $\alpha$ and $\beta$, to selectively boost local and summary tokens, respectively.

\subsection{Our Method}
\label{subsec:our_approach}

Our approach consists of two steps. First, we identify local and summary tokens using CLS-to-patch attention maps in the Vision Transformer (ViT). As discussed in the previous section, we select the top N\% of tokens with the highest CLS-to-patch attention in the top layer as local tokens, while the top N% in the bottom layer are selected as summary tokens. Then, we apply a two-step selective attention enhancement to $\mathbf{S_{\text{local}}}$ and $\mathbf{S_{\text{summary}}}$ with $\alpha$ and $\beta$, respectively. The values of $N$, $\alpha$, and $\beta$ are determined through experiments (Section~\ref{sec:ablation}).

\begin{equation}
    \mathbf{S_{\text{local}}} = \text{argmax}_N \left( \frac{1}{H} \sum_{h=1}^{H} \mathbf{A}_{\text{layer 1}}^h \right)
    \label{eq:local_tokens}
\end{equation}

\begin{equation}
    \mathbf{S_{\text{summary}}} = \text{argmax}_N \left( \frac{1}{H} \sum_{h=1}^{H} \mathbf{A}_{\text{layer 24}}^h \right)
    \label{eq:summary_tokens}
\end{equation}

\begin{equation}
    \bar{A}^{n,j} = \begin{cases}
        \bar{A}^{n,j} + \alpha \cdot |\bar{A}^{n,j}|, & \text{if } j \in \mathbf{S_{\text{local}}} \\
        \bar{A}^{n,j} + \beta \cdot |\bar{A}^{n,j}|, & \text{if } j \in \mathbf{S_{\text{summary}}} \\
    \end{cases}
    \label{eq:complete_attention}
\end{equation}

%% file: sec/4_exp.tex
\section{Evaluation}
\label{sec:experimental_setup}
%------------------------------------------------------------------------

\subsection{Setup}

We implement our approach on LLaVA-1.5~\cite{llava} in Huggingface Transformers framework~\cite{huggingface-transformer}. Our approach is designed to be easily extendable to other LVLMs including InstructBLIP \cite{dai2023instructblipgeneralpurposevisionlanguagemodels}, LLaMA-Adapter-v2 \cite{gao2023llama}, MiniGPT4 \cite{zhu2023minigpt}, Shikra \cite{chen2023shikra}, etc. since they share a similar architecture. We use an NVIDIA A100 (80GB) to run our experiments.

\subsection{Benchmarks and Metrics}
% Some common benchmark we will utilize eight benchmarks from the \emph{lmms-eval} evaluation framework~\cite{lmms-eval}, spanning three different task types: 

We sample 500 validation instances from \textbf{MSCOCO 2014} to assess hallucination. We use CHAIR\cite{rohrbach2018object}, POPE~\cite{pope} and instance-level Recall for hallucination detection. The CHAIR score evaluates hallucination in LVLMs by prompting the model to generate descriptions for input images followed by comparing these descriptions with the actual objects present. It consists of two metrics: instance-level hallucination $\text{CHAIR}_{I}$ and sentence-level hallucination $\text{CHAIR}_{S}$. 

\begin{align}
\text{CHAIR}_I &= \frac{\left|\{ \text{hallucinated objects} \} \right|}{\left|\{ \text{all mentioned objects} \} \right|} \\
\text{CHAIR}_S &= \frac{\left|\{ \text{captions with hallucinated objects} \} \right|}{\left|\{ \text{all captions} \} \right|}
\end{align}

POPE~\cite{pope}, on the other hand, assesses LVLMs by using binary prompts (e.g., “Is there a keyboard in this image?”) to evaluate object recognition accuracy, reporting accuracy, F1 score, and the proportion of "yes" responses, especially under adversarial conditions. The instance-level Recall score in our evaluation will assess whether the descriptions accurately capture the essential visual content of the image.

\begin{align}
\text{Recall} &= \frac{\left|\{ \text{non-hallucinated objects} \} \right|}{\left|\{ \text{all existing objects} \} \right|}
\end{align}

\subsection{Baselines and Results}
We compare our method with the baseline PAI~\cite{liu2024payingattentionimagetrainingfree}, a training-free approach that balances image comprehension and language inference by adaptively adjusting attention weights and subtracting text-only logits. As shown in Table~\ref{tab:main_results}, our method significantly reduces hallucination, achieving a \textbf{61.9\% decrease} in CHAIR$_S$ (46.2\% $\rightarrow$ 17.6\%) and a \textbf{71.0\% reduction} in CHAIR$_I$ (13.8\% $\rightarrow$ 4.0\%), outperforming PAI’s \textbf{46.8\%} and \textbf{50.7\%} reductions, respectively. Incorporating spatial token selection further lowers CHAIR$_S$ to \textbf{15.4\%} but slightly affects CHAIR$_I$ and F1-Score, highlighting a trade-off between hallucination reduction and generation accuracy. Despite a \textbf{moderate F1 decrease} (75.9\% $\rightarrow$ 71.8\%), our selective attention and head-specific modulation effectively mitigate hallucinations while preserving overall model performance along with offering flexibility.

\begin{table}[H]
    \centering
    \footnotesize
    \renewcommand{\arraystretch}{1.1} % Adjusts row spacing
    \setlength{\tabcolsep}{2pt} % Reduces column padding for compactness
    \caption{Comparison of hallucination metrics across different approaches on the MSCOCO dataset. CHAIR$_S$ and CHAIR$_I$ measure sentence-level and instance-level hallucination rates, respectively (lower is better). Our method achieves substantial reductions in hallucination while maintaining reasonable F1 scores.}
    \begin{tabular}{lccc}
        \toprule
        \textbf{Model} & \textbf{CHAIR$_S$} & \textbf{CHAIR$_I$} & \textbf{F1-Score} \\
        \midrule
        Original & 46.2 & 13.8 & 75.9 \\
        PAI & 24.6 & 6.8 & 74.7 \\
        \rowcolor{LightOrange} PAINT & 17.6 & 4.0 & 71.8 \\  % Light green
        % \rowcolor{cvprblue} PAINT (w/ Spatial) & 15.4 & 4.8 & 69.8 \\  % Light yellow
        \bottomrule
    \end{tabular}
    
    \label{tab:main_results}
\end{table}

\subsection{Ablation Studies}
\label{sec:ablation}

To better understand the contribution of different components in our approach, we conduct extensive ablation studies focusing on the impact of attention modification parameters. All experiments in this section use 25\% selected Local tokens without spatial token selection to isolate the effects of individual components.

% \subsubsection{Impact of $\alpha$}

\subsubsection{Impact of \texorpdfstring{$\alpha$}{alpha}} 
 We first investigate the effect of the local token attention amplification factor ($\alpha$). Table~\ref{tab:alpha_ablation} shows the performance metrics across different $\alpha$ values. The results demonstrate a clear pattern where moderate amplification (around $\alpha = 0.7$) achieves optimal performance. At this value, we observe a significant reduction in hallucination (CHAIR$_S$ = 18.0\%, CHAIR$_I$ = 3.9\%) while maintaining strong task performance (F1-score = 72.0\%).

Lower $\alpha$ values (0.1--0.5) show relatively consistent but suboptimal performance, with CHAIR$_S$ ranging from 47.4\% to 49.4\% and CHAIR$_I$ from 12.4\% to 13.3\%. However, increasing $\alpha$ beyond 0.7 leads to a sharp degradation in performance, with extreme values ($\alpha = 0.9$) resulting in very poor hallucination metrics (CHAIR$_S$ = 1.0\%, CHAIR$_I$ = 6.1\%) and significantly reduced F1-scores (8.7\%).

\begin{table}[H]
    \centering
   
        \footnotesize
        \captionof{table}{Effect of Different Local Token Attention Amplification Factors ($\alpha$) on Hallucination Reduction and Model Stability.}
        \label{tab:alpha_ablation}
        \setlength{\tabcolsep}{4pt}  % Reduce column spacing
        \renewcommand{\arraystretch}{0.9}  % Reduce row height
        \small
        \begin{tabular}{@{}lcccc@{}}
            \toprule
            $\alpha$ & CHAIR$_S$ & CHAIR$_I$ & F1 & Length \\
            \midrule
            0.1 & 47.4 & 12.7 & 77.5 & 96.9 \\
            0.3 & 47.2 & 12.4 & 77.0 & 96.0 \\
            0.4 & 48.2 & 12.9 & 76.8 & 97.3 \\
            0.5 & 49.4 & 13.3 & 77.0 & 99.9 \\
            % \rowcolor[HTML]{D5E8D4} \textbf{0.7}
            \rowcolor{LightOrange} \textbf{0.7}
            & \textbf{18.0} & \textbf{3.9} & \textbf{72.0} & \textbf{179.7} \\
            0.8 & 4.6 & 1.0 & 51.1 & 41.1 \\
            0.9 & 1.0 & 6.1 & 8.7 & 10.0 \\
            \bottomrule
        \end{tabular}
\end{table}

% \subsubsection{Impact of $\beta$}  
\subsubsection{Impact of \texorpdfstring{$\beta$}{beta}} 
We also examine the impact of the summary token attention factor ($\beta$). Table~\ref{tab:beta_ablation} presents results across different $\beta$ values. The optimal performance is achieved at $\beta = 0.4$, which balances hallucination reduction with overall model capability. This configuration yields CHAIR$_S$ = 17.6\% and CHAIR$_I$ = 4.0\%, with an F1-score of 71.8\%.

        The results show a consistent trend where increasing $\beta$ from 0.1 to 0.4 gradually improves performance across all metrics. However, beyond $\beta = 0.4$, the model becomes unstable, as indicated by the absence of meaningful results at $\beta = 0.5$. This suggests that while some amplification of summary token attention is beneficial, excessive emphasis on summary tokens can disrupt the model's ability to generate coherent outputs.
\begin{table}[H]

        \centering
        \footnotesize
        \captionof{table}{Ablation Study on the Role of Summary Token Attention Factor ($\beta$) in Controlling Hallucination and Model Stability.}
        \label{tab:beta_ablation}
        \setlength{\tabcolsep}{4pt}  % Reduce column spacing
        \renewcommand{\arraystretch}{0.9}  % Reduce row height
        \small
        \begin{tabular}{@{}lcccc@{}}
            \toprule
            $\beta$ & CHAIR$_S$ & CHAIR$_I$ & F1 & Length \\
            \midrule
            0.1 & 47.4 & 12.7 & 77.5 & 96.9 \\
            0.2 & 47.2 & 12.4 & 77.0 & 96.0 \\
            0.3 & 48.2 & 12.9 & 76.8 & 97.3 \\
            \rowcolor{LightOrange}
            
            \textbf{0.4} & \textbf{17.6} & 
            \textbf{4.0} & \textbf{71.8} & \textbf{179.7} \\
            0.5 & -- & -- & -- & -- \\  % Unstable values, indicated with dashes
            \bottomrule
        \end{tabular}
  
\end{table}

% \subsubsection{Impact of $N$}
\subsubsection{\texorpdfstring{Impact of $N$}{Impact of N}}

We analyze the effect of local token selection (Top Token Ratio) on hallucination metrics (CHAIR\textsubscript{S}, CHAIR\textsubscript{I}), F1 score, and output length (Table~\ref{tab:top_token_metrics}). The optimal setting is 25\%, where amplified attention reduces CHAIR\textsubscript{S} to 17.6, CHAIR\textsubscript{I} to 4.0, and maintains a balanced F1 = 71.8, though with an increased length of 187.8. Increasing to 30\% further lowers CHAIR\textsubscript{S} (12.8), but significantly drops F1 (-5.6\%). Lower ratios (10\%–20\%) achieve higher F1 (74.3–77.2) but retain high hallucination levels (CHAIR\textsubscript{S} = 44.0–29.6). Thus, 25\% provides the best trade-off between hallucination reduction and task performance.
\begin{table}[H]
    
        \centering
        \footnotesize
        \captionof{table}{Effect of Token Selection Ratio $N$: Top tokens are selected to analyze the impact on performance and model Stability. All experiments use $\alpha$ = 0.7, $
\beta$ = 0.4}
        \label{tab:top_token_metrics}
        \setlength{\tabcolsep}{6pt}  % Adjust column spacing
        \renewcommand{\arraystretch}{0.9}  % Reduce row height
        \small
        \begin{tabular}{@{}lcccc@{}}
            \toprule
            \textbf{Token ratio} & \textbf{CHAIR$_S$} & \textbf{CHAIR$_I$} & \textbf{F1} & \textbf{Length} \\
            \midrule
            0.1  & 44.0 & 12.0 & 77.2 & 99.4  \\
            0.15 & 38.2 & 11.3 & 77.2 & 101.2 \\
            0.2  & 29.6 & 9.2  & 74.3 & 113.0 \\
            \rowcolor{LightOrange} \textbf{0.25} & \textbf{17.6} & \textbf{4.0} & \textbf{71.8} & \textbf{187.8} \\
            0.3  & 12.8 & 4.7  & 67.8 & 246.3 \\
            \bottomrule
        \end{tabular}
    
\end{table}

%% file: sec/6_conclusion.tex
\section{Conclusion}
In this paper, we introduce a new method to reduce hallucinations in Large Vision-Language Models. We identified two types of visual tokens that transfer information from the vision encoder to the language model: local tokens, which provide detailed image features, and summary tokens, which capture overall image content. Unlike existing methods that boost all visual tokens equally, we showed that selectively enhancing local and summary tokens leads to better hallucination reduction. Our approach reduces hallucinations by 61.9\% at the sentence level and 71.0\% at the instance level. This method is simple, effective, and can be applied in a plug-and-play manner. We hope it inspires further research and real-world applications.